\documentclass[11pt,a4paper]{article}
\usepackage[hyperref]{naaclhlt2018}
\usepackage{times}
\usepackage{latexsym}
\usepackage{url}

\usepackage{caption}
\usepackage{float}
\usepackage{fixltx2e}

\newfloat{mixed}{thp}{lop}

\aclfinalcopy 
\def\aclpaperid{415} 

\usepackage{amsmath}
\usepackage{multirow}
\usepackage{booktabs}
\usepackage{pgfplots}

\usepackage{subcaption}
\usepackage{filecontents}

\usetikzlibrary{patterns}
\usetikzlibrary{calc,positioning}
\usepgfplotslibrary{groupplots}
\PassOptionsToPackage{monochrome}{xcolor}

\usepackage[compact]{titlesec}

\usepackage[final]{microtype}
\usepackage{xspace}
\newcommand{\mtwo}{M$^2$\xspace}

\pgfplotsset{
ticklabel style = {font=\small},
/tikz/mark size={2},
ticklabel style = {font=\small},
legend style = {font=\small},
every axis plot/.append style={thick},
y tick label style={/pgf/number format/.cd, fixed, fixed zerofill, precision=1, /tikz/.cd},
x tick label style={/pgf/number format/.cd, fixed, fixed zerofill, precision=1, /tikz/.cd},
 cycle list name=black white,
}

\definecolor{bblue}{HTML}{4F81BD}
\definecolor{rred}{HTML}{C0504D}
\definecolor{ggreen}{HTML}{9BBB59}
\definecolor{ppurple}{HTML}{9F4C7C}
\definecolor{oorange}{HTML}{F08000}

\title{Approaching Neural Grammatical Error Correction\\ as a Low-Resource Machine Translation Task}

\author{Marcin Junczys-Dowmunt \\
  Microsoft \\
  {\tt marcinjd@microsoft.com} \\ \And
  Roman Grundkiewicz \\
  University of Edinburgh \\
  {\tt rgrundki@inf.ed.ac.uk} \\ \AND
  Shubha Guha \\
  University of Edinburgh \\
  {\tt sguha@ed-alumni.net} \\ \And
  Kenneth Heafield \\
  University of Edinburgh \\
  {\tt kheafiel@inf.ed.ac.uk} \\}

\date{}

\begin{document}
\maketitle
\begin{abstract}
  Previously, neural methods in grammatical error correction (GEC) did not reach state-of-the-art results compared to phrase-based statistical machine translation (SMT) baselines. We demonstrate parallels between neural GEC and low-resource neural MT and successfully adapt several methods from low-resource MT to neural GEC.
  We further establish guidelines for trustable results in neural GEC and propose a set of model-independent methods for neural GEC that can be easily applied in most GEC settings.
  Proposed methods include adding source-side noise, domain-adaptation techniques, a GEC-specific training-objective, transfer learning with monolingual data, and ensembling of independently trained GEC models and language models.
  The combined effects of these methods result in better than state-of-the-art neural GEC models that outperform previously best neural GEC systems by more than 10\% M$^2$ on the CoNLL-2014 benchmark and 5.9\% on the JFLEG test set. Non-neural state-of-the-art systems are outperformed by more than 2\% on the CoNLL-2014 benchmark and by 4\% on JFLEG.
\end{abstract}

\section{Introduction}

Most successful approaches to automated grammatical error correction (GEC) are based on methods from statistical machine translation (SMT), especially the phrase-based variant. For the CoNLL 2014 benchmark on grammatical error correction \cite{ng2014conll}, \newcite{junczys2016phrase} established a set of methods for GEC by SMT that remain state-of-the-art. Systems \cite{chollampatt2017connecting, yannakoudakis2017neural} that improve on results by \newcite{junczys2016phrase} use their set-up as a back-bone for more complex systems.

The view that GEC can be approached as a machine translation problem by translating from erroneous to correct text originates from \newcite{brockett2006correcting} and resulted in many systems  \cite[e.g.][]{felice2014grammatical,susanto2014system} that represented the current state-of-the-art at the time.

In the field of machine translation proper, the emergence of neural sequence-to-sequence methods and their impressive results have lead to a paradigm shift away from phrase-based SMT towards neural machine translation (NMT). During WMT 2017 \cite{WMT:2017} authors of pure phrase-based systems offered ``unconditional surrender''\footnote{\newcite{ding-EtAl:2017:WMT} on their news translation shared task poster \url{http://www.cs.jhu.edu/~huda/papers/jhu-wmt-2017.pdf}} to NMT-based methods.

Based on these developments, one would expect to see a rise of state-of-the-art neural methods for GEC, but as \newcite{junczys2016phrase} already noted, this is not the case. Interestingly, even today, the top systems on established GEC benchmarks are still mostly phrase-based or hybrid systems \cite{chollampatt2017connecting, yannakoudakis2017neural,napoles2017systematically}. The best ``pure'' neural systems \cite{ji2017nested,sakaguchi2017grammatical,DBLP:conf/emnlp/SchmaltzKRS17} are several percent behind.\footnote{After submission of this work, \newcite{chollampatt2018mlconv} published impressive new results for neural GEC with some overlap with our methods. However, our results stay ahead on all benchmarks while using simpler models.}

\begin{figure}
 \centering
 \includegraphics[width=0.49\textwidth]{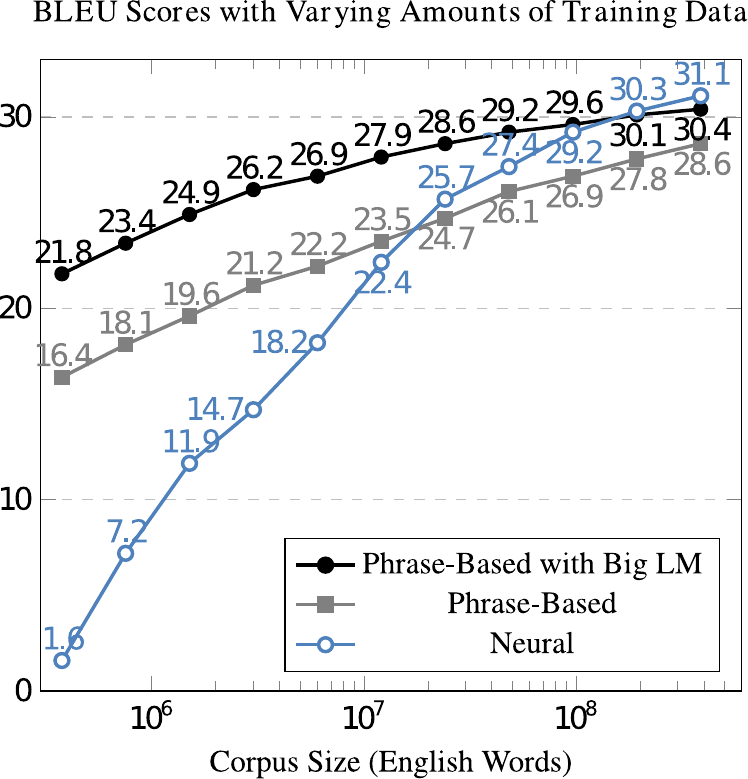}
 \caption{BLEU scores for English-Spanish systems trained on 0.4M to 385.7M words of parallel data. Source:
 Koehn and Knowles (2017)} 
 \label{koehnknowles}
\end{figure}

If we look at recent MT work with this in mind, we find one area where phrased-based SMT dominates over NMT: low-resource machine translation. \newcite{koehn2017six} analyze the behavior of NMT versus SMT for English-Spanish systems trained on 0.4 million to 385.7 million words of parallel data, illustrated in Figure~\ref{koehnknowles}. Quality for NMT starts low for small corpora, outperforms SMT at a corpus size of about 15 million words, and with increasing size beats SMT with a large in-domain language model.

Table~\ref{tab:data} lists existing training resources for the English as-a-second-language (ESL) grammatical error correction task. Publicly available resources, NUS Corpus of Learner English (NUCLE) by \newcite{dahlmeier2013building}, Lang-8 NAIST \cite{mizumoto2012effect} and CLC FCE \cite{yannakoudakis2011new} amount to about 27M tokens. 
Among these the Lang-8 corpus is quite noisy and of low quality.
The Cambridge Learner Corpus (CLC) by \newcite{nicholls2003cambridge} --- probably the best resource in this list --- is non-public and we would strongly discourage reporting results that include it as training data as this makes comparisons difficult.

\begin{table}[t]\centering
\begin{tabular}{lrrc}
\toprule
    Corpus              & Sent. & Tokens & Public \\
\midrule
    NUCLE*         &    57.1K &  1.2M & Yes \\
    Lang-8 NAIST*       & 1.9M & 25.0M & Yes \\
    CLC FCE & 30.9K & 0.5M & Yes \\
    CLC & 1.9M &  29.2M & No \\
\bottomrule
\end{tabular}
    \caption{Statistics for existing GEC training data sets. Data sets marked with * are used in this work.}
\label{tab:data}
\end{table}

Contrasting this with Fig.~\ref{koehnknowles}, we see that for about 20M tokens NMT systems start outperforming SMT models without additional large language models. Current state-of-the-art GEC systems based on SMT, however, all include large-scale in-domain language models either following the steps outlined in \newcite{junczys2016phrase} or directly re-using their domain-adapted Common-Crawl language model.

It seems that the current state of neural methods in GEC reflects the behavior for NMT systems trained on smaller data sets. Based on this, we conclude that we can think of GEC as a low-resource, or at most mid-resource, machine translation problem. This means that techniques proposed for low-resource (neural) MT should be applicable to improving neural GEC results.

In this work we show that adapting techniques from low-resource (neural) MT and SMT-based GEC methods allows neural GEC systems to catch up to and outperform SMT-based systems. We improve over the previously best-reported neural GEC system \cite{ji2017nested} on the CoNLL 2014 test set by more than 10\% M$^2$,
over a comparable pure SMT system by \newcite{junczys2016phrase} by 6\%, and outperform the state-of-the-art result of \newcite{chollampatt2017connecting} by 2\%. On the JFLEG data set, we report the currently best results, outperforming the previously best pure neural system \cite{sakaguchi2017grammatical} by 5.9\% GLEU and the best reported results \cite{chollampatt2017connecting} by 3\% GLEU.

In Section~\ref{baseline}, we describe our NMT-based baseline for GEC, and follow recommendations from the MT community for a trustable neural GEC system.
In Section~\ref{better}, we adapt neural models to make better use of sparse error-annotated data, transferring low-resource MT and GEC-specific SMT methods to neural GEC. This includes a novel training objective for GEC.
We investigate how to leverage monolingual data for neural GEC by transfer learning in Section~\ref{mono} and experiment with language model ensembling in Section~\ref{mono2}.
Section~\ref{deeper} explores deep NMT architectures.
In Section~\ref{toolbox}, we provide an overview of the experiments and how results relate to the JFLEG benchmark. We also recommend a model-independent toolbox for neural GEC.

\section{A trustable baseline for neural GEC}
\label{baseline}

In this section, we combine insights from \newcite{junczys2016phrase} for grammatical error correction by phrase-based statistical machine translation and from \newcite{denkowski17nmt} for trustable results in neural machine translation to propose a trustable baseline for neural grammatical error correction.

\subsection{Training and test data}

To make our results comparable to state-of-the-art results in the field of GEC, we limit our training data strictly to public resources. In the case of error-annotated data, as marked in Table~\ref{tab:data}, these are the NUCLE \cite{dahlmeier2013building} and Lang-8 NAIST \cite{mizumoto2012effect} data sets. We do not include the FCE corpus \cite{yannakoudakis2011new} to maintain comparability to \newcite{junczys2016phrase} and \newcite{chollampatt2017connecting}. We strongly urge the community to not use the non-public CLC corpus for training, unless contrastive results without this corpus are provided as well.

We choose the CoNLL-2014 shared task test set \cite{ng2014conll} as our main benchmark and the test set from the 2013 edition of the shared task \cite{ng2013conll} as a development set. For these benchmarks we report MaxMatch (M$^2$) scores \cite{dahlmeier2012better}.
Where appropriate, we will provide results on the JFLEG dev and test sets \cite{napoles2017jfleg} using the GLEU metric \cite{sakaguchi2016reassessing} to demonstrate the generality of our methods. Table~\ref{tab:teststats} summarizes test/dev set statistics for both tasks.

For most our experiments, we report M$^2$ on CoNLL-2013 test (Dev) and precision (Prec.), recall (Rec.),  M$^2$ (Test) on the CoNLL-2014 test set.

\begin{table}[t]\centering
\begin{tabular}{lrcc}
\toprule
    Test/Dev set         & Sent. & Annot. & Metric \\
\midrule
   CoNLL-2013 test    &     1,381 &     1  & M$^2$ \\
   CoNLL-2014 test    &     1,312 &     2  & M$^2$ \\
   JFLEG dev          &       754 &     4  & GLEU \\
   JFLEG test         &       747 &     4  & GLEU \\
\bottomrule
\end{tabular}
\caption{Statistics for test and development data.}
\label{tab:teststats}
\end{table}

\subsection{Preprocessing and sub-words}
\label{prepro}
As both benchmarks, CoNLL and JFLEG, are provided in NLTK-style tokenization \cite{Bird:2009:NLP:1717171}, we use the same tokenization scheme for our training data. We truecase line beginnings and escape special characters using scripts included with Moses \cite{koehn2007moses}. 
Following \newcite{sakaguchi2017grammatical}, we apply the Enchant\footnote{\url{https://github.com/AbiWord/enchant}} spell-checker to the JFLEG data before evaluation. No spell-checking is used for the CoNLL test sets.

We follow the recommendation by \newcite{denkowski17nmt} to use byte-pair encoding (BPE) sub-word units \cite{sennrich2016bpe} to solve the large-vocabulary problem of NMT. This is a well established procedure in neural machine translation and has been demonstrated to be generally superior to UNK-replacement methods. It has been largely ignored in the field of grammatical error correction even when word segmentation issues have been explored \cite{ji2017nested,DBLP:conf/emnlp/SchmaltzKRS17}. To our knowledge, this is the first work to use BPE sub-words for GEC, however, an analysis on advantages of word versus sub-word or character level segmentation is beyond the scope of this paper. A set of 50,000 monolingual BPE units is trained on the error-annotated data and we segment training and test/dev data accordingly. Segmentation is reversed before evaluation.

\subsection{Model and training procedure}
\label{sec.model}
Implementations of all models explored in this work\footnote{Models, system configurations and outputs are available from \url{https://github.com/grammatical/neural-naacl2018}} are available in the Marian\footnote{\url{ https://github.com/marian-nmt/marian}} toolkit \cite{mariannmt}. The attentional encoder-decoder model in Marian is a re-implementation of the NMT model in Nematus \cite{sennrich2017nematus}. The model differs from the model introduced by \newcite{bahdanau2014neural} by several aspects, the most important being the conditional GRU with attention for which \newcite{sennrich2017nematus} provide a concise description.

All embedding vectors consist of 512 units; the RNN states of 1024 units. The number of BPE segments determines the size of the vocabulary of our models, i.e.~50,000 entries. Source and target side use the same vocabulary.
To avoid overfitting, we use variational dropout \cite{NIPS2016_6241} over GRU steps and input embeddings with probability 0.2. We optimize with Adam \cite{kingma2014adam} with an average mini-batch size of ca.~200.
All models are trained until convergence (early-stopping with a patience of 10 based on development set cross-entropy cost), saving model checkpoints every 10,000 mini-batches. The best eight model checkpoints w.r.t.~the development set M$^2$ score of each training run are averaged element-wise \cite{junczys2016neural} resulting in a final single  model.
During decoding we use a beam-size of 24 and normalize model scores by length.\footnote{We used a larger beam-size than usual due to experiments with re-ranking of n-best lists not included in the paper. We did not see any differences compared to smaller beams.}

\begin{table}[t]\centering
 \begin{tabular}{ccccc|cc}\toprule
  & \multicolumn{4}{c|}{CoNLL} & \multicolumn{2}{c}{JFLEG} \\
Run & Dev & Prec. & Rec. & Test & Dev & Test \\ \midrule
1 & 20.2 & 68.6 & 11.8 & 34.9 & 47.6 & 52.3 \\
2 & 21.3 & 64.6 & 10.3 & 31.5 & 47.1 & 51.8 \\
3 & 21.7 & 64.8 & 10.6 & 32.0 & 47.1 & 52.4 \\
4 & 22.0 & 67.1 & 10.9 & 33.0 & 47.1 & 52.0\\ \midrule
Avg & 21.3 & -- & -- & 32.9 & 47.2 & 52.1 \\
Ens & 19.3 & 70.8 & 9.5 & 30.9 & 47.0 & 52.5\\ \bottomrule
 \end{tabular}

 \caption{Instable results for multiple baseline runs versus average and ensemble --- for the CoNLL benchmark.}
 \label{instable}
\end{table}

\subsection{Optimizer instability}

\newcite{junczys2016phrase} noticed that discriminative parameter tuning for GEC by phrase-based SMT leads to unstable M$^2$ results between tuning runs. This is a well-known effect for SMT parameter tuning and \newcite{DBLP:conf/acl/ClarkDLS11} recommend reporting results for multiple tuning runs. \newcite{junczys2016phrase} perform four tuning runs and calculate parameter centroids following \newcite{Cettolo:Mauro}.

Neural sequence-to-sequence training is discriminative optimization and as such prone to instability. We already try to alleviate this by averaging over eight best checkpoints, but as seen in Table~\ref{instable}, results for M$^2$ remain unstable for runs with differently initialized weights. An amplitude of 3 points M$^2$ on the CoNLL-2014 test set is larger than most improvements reported in recent papers. None of the recent works on neural GEC account for instability, hence it is unclear if observed outcomes are actual improvements or lucky picks among by-products of instability. We therefore strongly suggest to provide results for multiple independently trained models. Otherwise improvements of less than 2 or 3 points of M$^2$ remain doubtful.
Interestingly, GLEU on the JFLEG data seems to be more stable than M$^2$ on CoNLL data.

\subsection{Ensembling of independent models}

Running multiple experiments to provide averaged results seems prohibitively expensive, but
\newcite{denkowski17nmt} and others \cite[e.g.][]{Sutskever:2014:SSL:2969033.2969173,sennrich-EtAl:2017:WMT} show that ensembling of independently trained models leads to consistent rewards for MT. For our baseline in Table~\ref{instable} the opposite seems to be true for M$^2$. This is likely the reason why no other work on neural GEC mentions results for ensembles.

On closer inspection, however, we see that the drop in M$^2$ for ensembles is due to a precision bias. M$^2$ being an F-score penalizes increasing distance between precision and recall. The increase in precision for ensembles is to be expected and we see it later consistently for all experiments. Ensembles choose corrections for which all independent models are fairly confident. This leads to fewer but better corrections, hence an increase in precision and a drop in recall. If the models are weak as our baseline, this can result in a lower score. It would, however, be unwise to dismiss ensembles, as we can use their bias towards precision to our advantage whenever they are combined with methods that aim to increase recall. This is true for nearly all remaining experiments.

\section{Adaptations for GEC}
\label{better}

\begin{table}[t]

\centering
\begin{tikzpicture}
\begin{axis}[
ylabel=M\textsuperscript{2},
every axis y label/.style={
     at={(ticklabel* cs:0.93)},
     anchor=south east, align=right
},
height=0.2\textheight,
width=0.8\linewidth,
scale only axis,
enlarge y limits,
ymajorgrids, xmajorgrids,
major grid style={dotted},
xtick={1,2,3,4,5,6},
ytick={40,42,...,50},
legend cell align=left,
legend style={column sep=10pt},
xticklabels={NMT-Baseline,+Dropout-Src,+Domain-Adapt.,+Error-Adapt.,+Tied-Emb.,+Edit-MLE},
xmin=1.2,
xmax=6.8,
legend pos=south east,
xticklabel style={align=right, rotate=45, anchor=north east},
every node near coord/.append style={anchor=south, font=\small, /pgf/number format/.cd, fixed, fixed zerofill, precision=1, /tikz/.cd},
]
\addplot+[solid, mark=-,
mark options={solid,fill=black,black},
gray, error bars/.cd,y dir=both,y explicit, error bar style={solid, black},
    error mark options={
      rotate=90,
      black,
      mark size=6pt,
      line width=1pt
    }]
table[x index=0,
      y expr={(\thisrow{m1}+\thisrow{m2}+\thisrow{m3}+\thisrow{m4})/4},
      y error plus expr={max(\thisrow{m1},\thisrow{m2},\thisrow{m3},\thisrow{m4})-(\thisrow{m1}+\thisrow{m2}+\thisrow{m3}+\thisrow{m4})/4},
      y error minus expr={-min(\thisrow{m1},\thisrow{m2},\thisrow{m3},\thisrow{m4})+(\thisrow{m1}+\thisrow{m2}+\thisrow{m3}+\thisrow{m4})/4},
      ] {test2014.simple};

\addplot+[nodes near coords, solid, mark=*, bblue,mark options={fill=bblue}]
table[x index=0, y=ens] {test2014.simple};

\legend{Average of 4, Ensemble of 4}
\end{axis}
\end{tikzpicture}
 \begin{tabular}{lccccc} \toprule
 Model & Dev & Prec. & Rec. & Test \\ \midrule
 Baseline       & 19.3 & 70.8 & 9.5 & 30.9 \\
 +Dropout-Src.  & 27.5 & 72.4 & 15.5 & 41.7 \\
 +Domain-Adapt. & 30.0 & 69.2 & 17.3 & 43.3 \\
 +Error-Adapt.  & 34.5 & 70.8 & 20.8 & 47.8 \\
 +Tied-Emb.     & 33.0 & 73.0 & 20.2 & 48.0 \\
 +Edit-MLE      & 37.6 & 65.3  & 27.1 & 51.0 \\ \bottomrule
 \end{tabular}
 \captionof{table}{Results (\mtwo) on the CoNLL benchmark for GEC-specific adaptations.}
\label{fig:adapt}
\end{table}

The methods described in this section turn our baseline into a more GEC-specific system. Most have been inspired by techniques from low-resource MT or closely related domain-adaptation techniques for NMT. All modifications are applied incrementally, later models include enhancements from the previous ones.

\subsection{Source-word dropout as corruption}

GEC can be treated as a denoising task where grammatical errors are corruptions that have to be reduced. By introducing more corruption on the source side during training we can teach the model to reduce trust into the source input and to apply corrections more freely. Dropout is one way to introduce noise, but for now we only drop out single units in the embedding or GRU layers, something the model can easily recover from. To make the task harder, we add dropout over  source words, setting the full embedding vector for a source word to $1/p_{\textrm{src}}$ with a probability of $p_{\textrm{src}}$. During our experiments, we found $p_{\textrm{src}}=0.2$ to work best.

Table~\ref{fig:adapt} show impressive gains for this simple method (+Dropout-Src.). Results for the ensemble match the previously best results on the CoNLL-2014 test set for pure neural systems (without the use of an additional monolingual language model) by \newcite{ji2017nested} and \newcite{DBLP:conf/emnlp/SchmaltzKRS17}.

\subsection{Domain adaptation}
The NUCLE corpus matches the domain of the CoNLL benchmarks perfectly. It is however much smaller than the Lang-8 corpus. A setting like this seems to be a good fit for domain-adaptation techniques.  \newcite{sennrich-haddow-birch:2016:WMT} oversample in-domain news data in a larger non-news training corpus. We do the same by adding the NUCLE corpus ten times to the training corpus. This can also be seen as similar to \newcite{junczys2016phrase} who tune phrase-based SMT parameters on the entire NUCLE corpus.
Respectable improvements on both CoNLL test sets (+Domain-Adapt. in Table~\ref{fig:adapt}) are achieved.

\subsection{Error adaptation}
\label{selection}
\newcite{junczys2016phrase} noticed that when tuning on the entire NUCLE corpus, even better results can be achieved if the error rate of NUCLE is adapted to the error rate of the original dev set. In NUCLE only 6\% of tokens contain errors, while the CoNLL-2013 test set has an error-rate of about 15\%. Following \newcite{junczys2016phrase}, we remove correct sentences from the ten-fold oversampled NUCLE data greedily until an error-rate of 15\% is achieved. This can be interpreted as a type of GEC-specific domain adaptation.
We mark this method as +Domain-Adapt.~in Table~\ref{fig:adapt} and report for the ensemble the so far strongest results for any neural GEC system on the CoNLL benchmark.

\subsection{Tied embeddings}

\newcite{DBLP:journals/corr/PressW16} showed that parameter tying between input and output embeddings\footnote{Output embeddings are encoded in the last output layer of a neural language or translation model.} for language models leads to improved perplexity. Similarly, three-way weight-tying between source, target and output embeddings for neural machine translation seems to improve translation quality in terms of BLEU while also significantly decreasing the number of parameters in the model.  In monolingual cases like GEC, where source and target vocabularies are (mostly) equal, embedding-tying seems to arise naturally. Output layer, decoder and encoder embeddings all share information which may further enhance the signal from corrective edits.
The M$^2$ scores for +Tied-Emb.~in Table~\ref{fig:adapt} are inconclusive, but we see improvements in conjunction with later modifications.

\subsection{Edit-weighted MLE objective}
\label{objective}

\begin{table}[t]\centering
 \begin{tabular}{ccccc|cc}\toprule
  & \multicolumn{4}{c|}{CoNLL} & \multicolumn{2}{c}{JFLEG} \\
  $\Lambda$ & Dev & Prec. & Rec. & Test & Dev & Test \\ \midrule
  1 & 33.5 & 67.5 & 20.8 & 46.6 & 48.9 & 53.9 \\ 
  3 & 36.8 & 59.8 & 28.8 & 49.2 & 51.2 & 56.5 \\ 
  5 & 36.2 & 54.0 & 30.8 & 47.0 & 50.9 & 55.7 \\ 
 \bottomrule
  \end{tabular}
 \caption{Results for model type +Tied-Emb. trained with edit-weighted MLE and chosen $\Lambda$.}\label{lambdas}
\end{table}

Previously, we applied error-rate adaptation to strengthen the signal from corrective edits in the training data. In this section, we investigate the effects of directly modifying the training loss to incorporate weights for corrective edits.

\begin{figure*}[ht]
\centering
\includegraphics[width=0.8\textwidth]{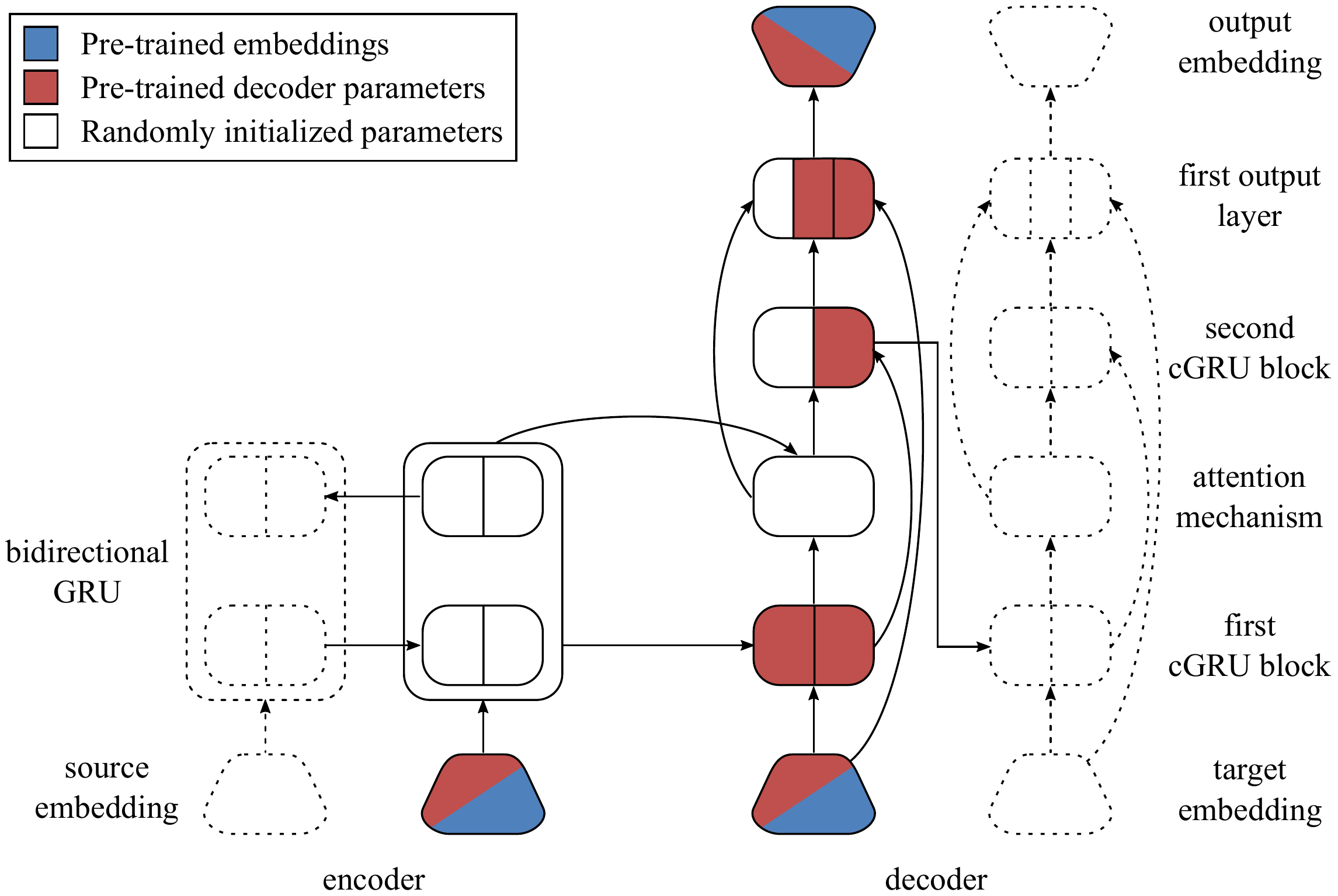}
\caption{Parameters pretrained on monolingual data are marked with colors. Blue indicates pre-trained embeddings with word2vec, red parameters have been pre-trained with the GRU-based language model only. All embedding layers have tied parameters.}\label{fig.pre}
\end{figure*}

Assuming that each target token $y_j$ has been generated by a source token $x_i$,
we scale the loss for each target token $y_j$ by a factor $\Lambda$ if $y_j$ differs from $x_i$, i.e.~if $y_j$ is part of an edit. Hence, log-likelihood loss takes the following form:
$$ L(x,y,a) = -\sum_{t=1}^{T_y} \lambda(x_{a_t}, y_t) \log P(y_t|x,y_{<t}), $$
$$ \lambda(x_{a_t}, y_t) = \left\{\begin{array}{cl} \Lambda & \mathrm{if}\; x_{a_t} \neq y_t \\ 1 & \mathrm{otherwise} \\ \end{array} \right. ,$$
where $(x, y)$ is a training sentence pair and $a$ is a word alignment $a_t \in \{0 ,1,\ldots,T_x\}$ such that source token $x_{a_t}$ generates target token $y_t$.
Alignments are computed for each sentence pair with fast-align \cite{DBLP:conf/naacl/DyerCS13}.

This is comparable to reinforcement learning towards GLEU as introduced by \newcite{sakaguchi2017grammatical} or training against diffs by \newcite{DBLP:conf/emnlp/SchmaltzKRS17}. In combination with previous modifications, edit-weighted Maximum Likelihood Estimation (MLE) weighting seem to outperform both methods. The parameter $\Lambda$ introduces an additional hyper-parameter that requires tuning for specific tasks and affects the precision/recall trade-off. Table~\ref{lambdas} shows $\Lambda=3$ seems to work best among the tested values when chosen to maximize M$^2$ on the CoNLL-2013 dev set.

For this setting, we achieve our strongest results of 50.95 M$^2$ on the CoNLL benchmark (system +Edit-MLE) yet. This outperforms the results of a phrase-based SMT system with a large domain-adapted language model from \newcite{junczys2016phrase} by 1\% M$^2$ and is the first neural system to beat this strong SMT baseline. 

\section{Transfer learning for GEC}
\label{mono}

Many ideas in low-resource neural MT are rooted in transfer learning. In general, one first trains a neural model on high-resource data and then uses the resulting parameters to initialize parameters of a new model meant to be trained on low-resource data only. Various settings are possible, e.g.~initializing from models trained on large out-of-domain data and continuing on in-domain data \cite{micelibarone-EtAl:2017:EMNLP2017} or using related language pairs \cite{DBLP:conf/emnlp/ZophYMK16}. Models can also be partially initialized by pre-training monolingual language models \cite{DBLP:conf/emnlp/RamachandranLL17} or only word-embeddings \cite{DBLP:conf/clic-it/GangiF17}. In GEC, \newcite{yannakoudakis2017neural} apply pre-trained monolingual word-embeddings as initializations for error-detection models to re-rank SMT n-best lists.
Approaches based on pre-training with monolingual data appear to be particularly well-suited to the GEC task.
\newcite{junczys2016phrase} published 300GB of compressed monolingual data used in their work to create a large domain-adapted Common-Crawl n-gram language model.\footnote{\url{https://github.com/grammatical/baselines-emnlp2016}}
We use the first 100M lines. Preprocessing follows section \ref{prepro} including BPE segmentation.

\begin{table}[t]

\begin{tikzpicture}
\begin{groupplot}[
group style={group size= 2 by 1, horizontal sep=2mm},
every axis y label/.style={
     at={(ticklabel* cs:0.93)},
     anchor=south east, align=right
},
ymin=44,ymax=54,
height=0.2\textheight,
width=0.4\linewidth,
scale only axis,
enlarge y limits,
ymajorgrids, xmajorgrids,
major grid style={dotted},
legend cell align=left,
legend style={column sep=10pt},
legend pos=south east,
xticklabel style={align=right, rotate=45, anchor=north east},
every node near coord/.append style={anchor=south, font=\small, /pgf/number format/.cd, fixed, fixed zerofill, precision=1, /tikz/.cd},
]

\nextgroupplot[ylabel=M\textsuperscript{2},xmin=0.2,xmax=3.8,
xticklabels={+Tied-Emb., +Pretrain-Emb., +Pretrain-Dec.},
xtick={1,2,3}]
\addplot+[solid, mark=-, gray,
mark options={solid,fill=black,black},
error bars/.cd,y dir=both,y 
explicit, 
error bar style={solid, black},
    error mark options={
      rotate=90,
      black,
      mark size=6pt,
    }]
table[x index=0,
      y expr={(\thisrow{m1}+\thisrow{m2}+\thisrow{m3}+\thisrow{m4})/4},
      y error plus expr={max(\thisrow{m1},\thisrow{m2},\thisrow{m3},\thisrow{m4})-(\thisrow{m1}+\thisrow{m2}+\thisrow{m3}+\thisrow{m4})/4},
      y error minus expr={-min(\thisrow{m1},\thisrow{m2},\thisrow{m3},\thisrow{m4})+(\thisrow{m1}+\thisrow{m2}+\thisrow{m3}+\thisrow{m4})/4},
      ] {test2014.simple1};

\addplot+[nodes near coords, solid, bblue, mark=*, mark options={fill=bblue}]
table[x index=0, y=ens] {test2014.simple1};

\nextgroupplot[xmin=0.2,xmax=3.8,yticklabels=none,
xtick={1,2,3},
xticklabels={+Edit-MLE, +Pretrain-Emb., +Pretrain-Dec.},
]
\addplot+[solid, mark=-, gray, 
mark options={solid,fill=black,black},
error bars/.cd,y dir=both,y explicit, error bar style={solid, black},
    error mark options={
      rotate=90,
      mark size=6pt,
    }]
table[x index=0,
      y expr={(\thisrow{m1}+\thisrow{m2}+\thisrow{m3}+\thisrow{m4})/4},
      y error plus expr={max(\thisrow{m1},\thisrow{m2},\thisrow{m3},\thisrow{m4})-(\thisrow{m1}+\thisrow{m2}+\thisrow{m3}+\thisrow{m4})/4},
      y error minus expr={-min(\thisrow{m1},\thisrow{m2},\thisrow{m3},\thisrow{m4})+(\thisrow{m1}+\thisrow{m2}+\thisrow{m3}+\thisrow{m4})/4},
      ] {test2014.simple2};

\addplot+[nodes near coords, solid, mark=*, bblue, mark options={fill=bblue}]
table[x index=0, y=ens] {test2014.simple2};

\legend{Average of 4, Ensemble of 4}
\end{groupplot}
\end{tikzpicture}
\centering
 \begin{tabular}{lccccc}\toprule
 Model & Dev & Prec. & Rec. & Test \\ \midrule
+Tied-Emb. & 33.0  & 73.0 & 20.2 & 48.0 \\       
$\;$ +Pretrain-Emb. & 35.5 & 69.1 & 22.8 & 49.1 \\ 
$\;$ +Pretrain-Dec. & 36.2 & 69.1 & 23.8 & 50.1 \\ 
\midrule
+Edit-MLE              & 37.6 & 65.3 & 27.1 & 51.0 \\ 
$\;$ +Pretrain-Emb. & 38.2 & 64.4 & 28.4 & 51.4 \\ 
$\;$ +Pretrain-Dec. & 40.3 & 65.2 & 32.2 & 54.1 \\ 
 \bottomrule
  \end{tabular}
  \caption{Results (\mtwo) on the CoNLL benchmark set for GEC-specific adaptations. 
}
\label{fig:transfer}
\end{table}

\subsection{Pre-training embeddings}
\label{pretrain}
Similarly to \newcite{DBLP:conf/clic-it/GangiF17} or \newcite{yannakoudakis2017neural}, we use Word2vec \cite{DBLP:journals/corr/abs-1301-3781} with standard settings to create word vectors. Since weights between source, target and output embeddings are tied, these embeddings are inserted once into the model, but affect computations three-fold, see the blue elements in Figure~\ref{fig.pre}. The remaining parameters of the model are initialized randomly. We refer to this adaptation as +Pretrain-Emb.

\subsection{Pre-training decoder parameters}

Following \newcite{DBLP:conf/emnlp/RamachandranLL17}, we first train a GRU-based language model on the monolingual data. The architecture of the language model corresponds as much as possible to the structure of the decoder of the  sequence-to-sequence model. All pieces that rely on the attention mechanism or the encoder have been removed. After training for two epochs, all red parameters  (including embedding layers) in Figure~\ref{fig.pre} are copied from the language model to the decoder. Remaining parameters are initialized randomly. This configuration is called +Pretrain-Dec. We pretrain each model separately to make sure that all weights have been initialized randomly. 

\subsection{Results for transfer learning}

Table~\ref{fig:transfer} summarizes the results for our transfer learning experiments. We compare the effects of pre-training with and without the edit-weighted MLE objective and can see that pre-training has significantly positive effects in both settings. 

The last result of 53.3\% M$^2$ on the CoNLL-2014 benchmark matches the currently highest reported numbers (53.14\% M$^2$) by \newcite{chollampatt2017connecting} for a much more complex system and outperforms the highest neural GEC system \cite{ji2017nested} by 8\% M$^2$.

\begin{table}[t]\centering
\begin{tabular}{lcccc}\toprule
  Model & Dev & Prec. & Rec. & Test \\ \midrule
 +Tied-Emb      & 33.0 & 73.0 & 20.2 & 48.0 \\ 
 +GRU-LM        & 40.2 & 59.8 & 36.2 & 52.9 \\ \midrule
 +Edit-MLE      & 37.6 & 65.3 & 27.1 & 51.0 \\ 
 +GRU-LM        & 40.3 & 61.9 & 34.5 & 53.4 \\ \midrule
+Pretrain-Dec. & 40.3 & 65.2 & 32.2 & 54.1 \\
 +GRU-LM        & 41.6 & 62.2 & 36.6 & 54.6 \\ 
  \bottomrule
\end{tabular}
\caption{Ensembling with a neural language model.}\label{lm}
\end{table}

\section{Ensembling with language models}
\label{mono2}
Phrase-based SMT systems benefit naturally from large monolingual language models, also in the case of GEC as shown by \newcite{junczys2016phrase}.
Previous work \cite{xie2016neural,ji2017nested} on neural GEC used n-gram language models to incorporate monolingual data. \newcite{xie2016neural} built a large 5-gram model and integrated it directly into their beam search algorithm, while \newcite{ji2017nested} re-use the language model provided by \newcite{junczys2016phrase} for n-best list re-ranking.

We already combined monolingual data with our GEC models via pre-training, but exploiting separate language models is attractive as no additional training is required. Here, we reuse the neural language model created for pre-training.

Similarly to \newcite{xie2016neural}, the score $s(y|x)$ for a correction $y$ of sentence $x$ is calculated as
 $$ s(y|x) = \frac{1}{|y|} \left[\sum_{i=1}^{4} \log P_i(y|x) + \alpha \log P_{\textrm{LM}}(y) \right] ,$$ where $P_i(y|x)$ is a translation probability for the $i$-th model in an ensemble of $4$. $P_\textrm{LM}(y)$ is the  language model probability for $y$ weighted by $\alpha$. We normalize by sentence length $|y|$.  Using the dev set, we choose $\alpha$ that maximizes this score via linear search in range $[0, 2]$ with step $0.1$.

Table~\ref{lm} summarizes results for language model ensembling with three of our intermediate configurations. All configurations benefit from the language model in the ensemble, although gains for the pre-trained model are rather small. 

\section{Deeper NMT models}
\label{deeper}

So far we analyzed model-independent\footnote{The pre-training procedure however needs to be adapted to model architecture if we want to take advantage of every shared parameter, otherwise matching parameter subsets could probably be used successfully.} methods --- only training data, hyper-parameters, parameter initialization, and the objective function were modified. In this section we investigate if these techniques can be generalized to deeper or different architectures. 

\subsection{Architectures}

We consider two state-of-the-art NMT architectures implemented in Marian:

\paragraph{Deep RNN} A deep RNN-based model \cite{micelibarone2017} proposed by \newcite{sennrich-EtAl:2017:WMT} for their WMT 2017 submissions. 
This model is based on the shallow model we used until now. It has single layer RNNs in the encoder and decoder, but increases depth by stacking multiple GRU-style blocks inside one RNN cell. A single RNN step passes through all blocks before recursion. The encoder RNN contains 4 stacked GRU blocks, the decoder 8 (1 + 7 due to the conditional GRU). Following \newcite{sennrich-EtAl:2017:WMT}, we enable layer-normalization in the RNN-layers. State and embedding dimensions used throughout this work and in \newcite{sennrich-EtAl:2017:WMT} are the same. 

\paragraph{Transformer} The self-attention-based model by \newcite{NIPS2017_7181}. We base our model on their default architecture of 6 complex attention/self-attention blocks in the encoder and decoder and use the same model dimensions --- embeddings vector size is 512 (as before), filter size is 2048. 

\subsection{Training settings}

As the deep models are less reliably trained with asynchronous SGD, we change the training algorithm to synchronous SGD and for both models follow the recipe proposed in \newcite{NIPS2017_7181}, with an effective base learning rate of 0.0003, learning rate warm-up during the first 16,000 iterations, and an inverse square-root decay after the warm-up. As before, we average the best 8 checkpoints. We increase dropout probability over RNN layers to 0.3 for Deep-RNN and similarly set dropout between transformer layers to 0.3. Source-word dropout as a noising technique remains unchanged. 

\subsection{Pre-training deep models}
We reuse all methods included up to +Pretrain-Dec. The pre-training procedure as described in section \ref{pretrain} needs to be modified in order to maximize the number of pre-trained parameters for the larger model architectures. Again, we train decoder-only models as typical language models by removing all elements that depend on the encoder, including attention-mechanisms over the source context. We can keep the decoder self-attention layers in the transformer model. We train for two epochs on our monolingual data reusing the hyper-parameters for the parallel case above.

\subsection{Results}
\begin{table}[t]\centering
\begin{tabular}{lcccc}\toprule
  Model & Dev & Prec. & Rec. & Test \\ \midrule
 +Pretrain-Dec. & 40.3 & 65.2 & 32.2 & 54.1 \\
 +GRU-LM        & 41.6 & 62.2 & 36.6 & 54.6 \\ \midrule
 +Deep-RNN & 41.1 & 64.3 & 35.2 & 55.2 \\
 +Deep-RNN-LM    & 41.9 & 61.3 & 40.2 & 55.5 \\ \midrule 
 +Transformer   & 41.5 & 63.0 & 38.9 & 56.1 \\
 +Transformer-LM & 42.9 & 61.9 & 40.2 & 55.8 \\ 
 \bottomrule
\end{tabular}
\caption{Shallow (Pretrain-Dec.)~versus deep ensembles, with and without corresponding language models.}\label{deep}
\end{table}

Table~\ref{deep} summarizes the results for deeper models on the CoNLL dev and test set. Both deep models improve significantly over the shallow model with the transformer model reaching our best result reported on the CoNLL 2014 test set. For that test set it seems that ensembling with language models that were used for pre-training is ineffective when measured with M$^2$; while on the JFLEG data measured with GLEU we see strong improvements (Fig. \ref{gleu}).

\section{A standard tool set for neural GEC}
\label{toolbox}

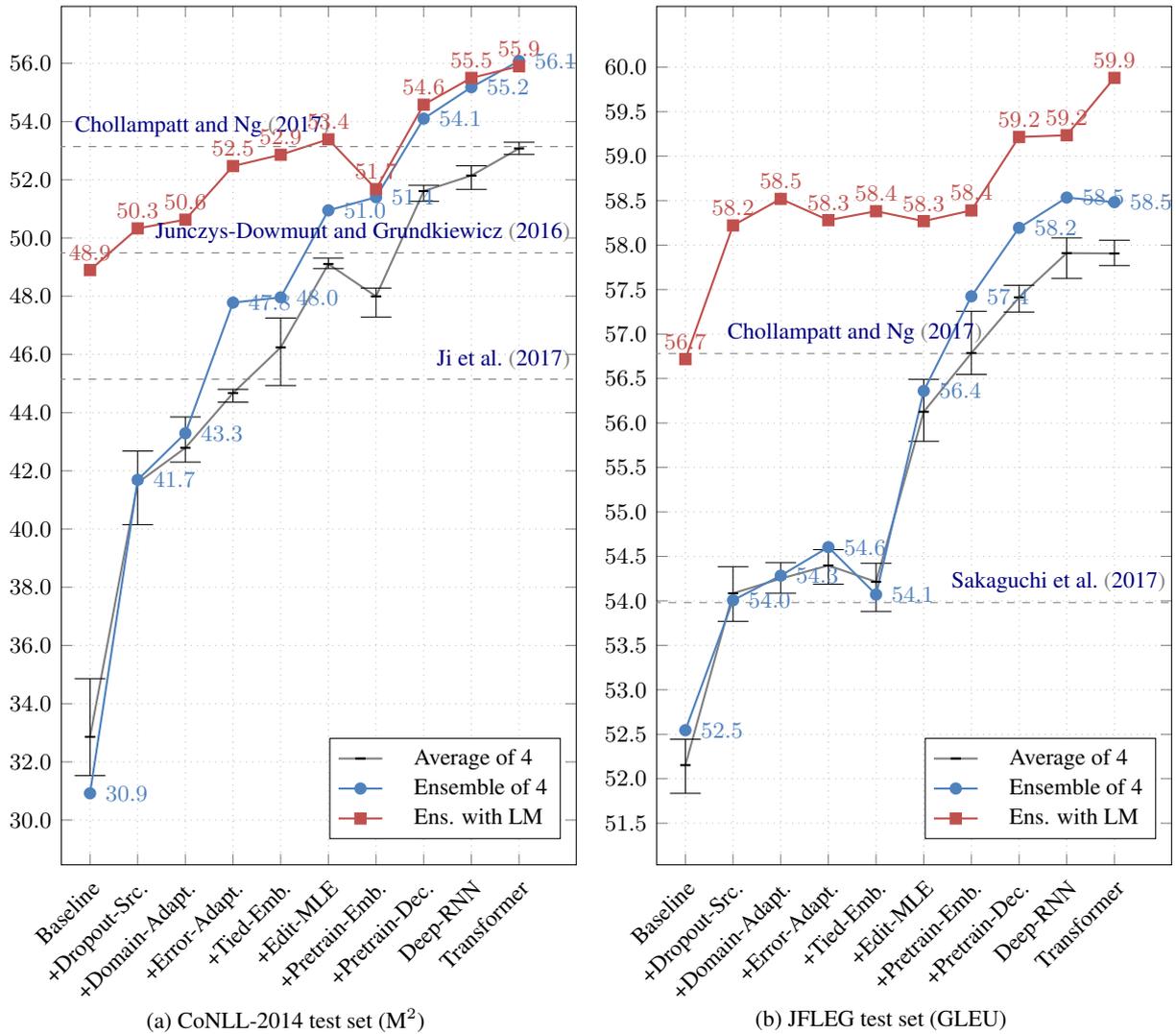
\begin{figure*}[t]

\begin{subfigure}[t]{0.49\textwidth}
\begin{tikzpicture}
\begin{axis}[
height=0.48\textheight,
width=0.9\linewidth,
scale only axis,
enlarge y limits,
ymajorgrids, xmajorgrids,
major grid style={dotted},
xtick={1,2,3,4,5,6,7,8,9,10},
legend cell align=left,
legend style={column sep=10pt},
every node near coord/.append style={anchor=south, font=\small, xshift=0pt, /pgf/number format/.cd, fixed, fixed zerofill, precision=1, /tikz/.cd},
xticklabels={Baseline,+Dropout-Src.,+Domain-Adapt.,+Error-Adapt.,
+Tied-Emb.,+Edit-MLE,+Pretrain-Emb.,+Pretrain-Dec.,Deep-RNN,Transformer},
ymax=55.5,
xmin=0.4,
xmax=11.2,
legend pos=south east,
xticklabel style={align=right, rotate=45, anchor=north east},
]
\addplot+[solid, gray, mark=-, 
mark options={solid, fill=black, black},
error bars/.cd,y dir=both,y explicit, error bar style={solid, black},
    error mark options={
      rotate=90,
      mark size=6pt
    }]
table[x index=0,
      y expr={(\thisrow{m1}+\thisrow{m2}+\thisrow{m3}+\thisrow{m4})/4},
      y error plus expr={max(\thisrow{m1},\thisrow{m2},\thisrow{m3},\thisrow{m4})-(\thisrow{m1}+\thisrow{m2}+\thisrow{m3}+\thisrow{m4})/4},
      y error minus expr={-min(\thisrow{m1},\thisrow{m2},\thisrow{m3},\thisrow{m4})+(\thisrow{m1}+\thisrow{m2}+\thisrow{m3}+\thisrow{m4})/4},
      ] {test2014};

\addplot+[nodes near coords,
every node near coord/.append style={anchor=west, xshift=2pt},
solid,  bblue, mark=*,mark options={solid, black, fill=bblue,bblue}]
table[x index=0, y=ens] {test2014};

\addplot+[solid, rred, nodes near coords, mark=square*, mark options={solid,fill=rred,rred}]
table[x index=0, y=lm] {test2014};

 \draw[dashed, gray] (axis cs:0,53.14) -- (axis cs:12,53.14)
 node[below, anchor=south west, pos=0.04] {\small \newcite{chollampatt2017connecting}};
 \draw[dashed, gray] (axis cs:0,49.49) -- (axis cs:12,49.49)
 node[below, anchor=south east, pos=0.94] {\small \newcite{junczys2016phrase}};
 
 \draw[dashed, gray] (axis cs:0,45.15) -- (axis cs:12,45.15)
 node[anchor=south east, pos=0.94] {\small \newcite{ji2017nested}};

\legend{Average of 4, Ensemble of 4, Ens. with LM}
\end{axis}
\end{tikzpicture}
\vspace{-8mm}\caption{CoNLL-2014 test set (M$^2$)}
\end{subfigure}\hfill
\begin{subfigure}[t]{0.49\textwidth}
\begin{tikzpicture}
\begin{axis}[
height=0.48\textheight,
width=0.9\linewidth,
scale only axis,
enlarge y limits,
ymajorgrids, xmajorgrids,
major grid style={dotted},
xtick={1,2,3,4,5,6,7,8,9,10},
ytick={51,51.5,...,60.1},
legend cell align=left,
legend style={column sep=10pt},
every node near coord/.append style={anchor=south, font=\small, xshift=0pt, /pgf/number format/.cd, fixed, fixed zerofill, precision=1, /tikz/.cd},
xticklabels={Baseline,+Dropout-Src.,+Domain-Adapt.,
+Error-Adapt.,+Tied-Emb.,+Edit-MLE,+Pretrain-Emb.,+Pretrain-Dec.,Deep-RNN,Transformer},
xmin=0.4,
xmax=11.2,
legend pos=south east,
xticklabel style={align=right, rotate=45, anchor=north east},
]
\addplot+[solid, gray, mark=-, 
    mark options={solid, black, fill=black},
error bars/.cd,y dir=both,y explicit, error bar style={solid, black},
    error mark options={
      solid,
      rotate=90,
      mark size=6pt
    }]
table[x index=0,
      y expr={(\thisrow{m1}+\thisrow{m2}+\thisrow{m3}+\thisrow{m4})/4},
      y error plus expr={max(\thisrow{m1},\thisrow{m2},\thisrow{m3},\thisrow{m4})-(\thisrow{m1}+\thisrow{m2}+\thisrow{m3}+\thisrow{m4})/4},
      y error minus expr={-min(\thisrow{m1},\thisrow{m2},\thisrow{m3},\thisrow{m4})+(\thisrow{m1}+\thisrow{m2}+\thisrow{m3}+\thisrow{m4})/4},
      ] {jflegtest};

\addplot+[nodes near coords,
every node near coord/.append style={anchor=west, xshift=2pt},
solid, bblue,  mark=*, mark options={solid, bblue,fill=bblue}]
table[x index=0, y=ens] {jflegtest};

\addplot+[nodes near coords, solid, rred,  mark=square*, mark options={solid, rred,fill=rred}]
table[x index=0, y=lm] {jflegtest};

 \draw[dashed, gray] (axis cs:0,56.78) -- (axis cs:12,56.78)
 node[below, anchor=south west, pos=0.14] {\small \newcite{chollampatt2017connecting}};
 
%
 \draw[dashed, gray] (axis cs:0,53.98) -- (axis cs:12,53.98)
 node[anchor=south east, pos=0.94] {\small \newcite{sakaguchi2017grammatical}};

\legend{Average of 4, Ensemble of 4, Ens. with LM}
\end{axis}
\end{tikzpicture}
\vspace{-8mm}
\caption{JFLEG test set (GLEU)\label{gleu}}\vspace{2mm}
\end{subfigure}
\caption{Comparison on the CoNLL-2014 test set and JFLEG test for all investigated methods.}\label{fig.all}
\end{figure*}

We summarize the results for our experiments in Figure~\ref{fig.all} and provide results on the JFLEG test set. Weights for the independent language model in the full ensemble were chosen on the respective dev sets for both tasks.
Comparing results according to both benchmarks and evaluation metrics (M$^2$ for CoNLL, GLEU for JFLEG), it seems we can isolate the following set of reliable methods for state-of-the-art neural grammatical error correction:

\begin{itemize}
 \itemsep0em
 \item Ensembling neural GEC models with monolingual language models;
 \item Dropping out entire source embeddings;
 \item Weighting edits in the training objective during optimization (+Edit-MLE);
 \item Pre-training on monolingual data;
 \item Ensembling of independently trained models;
 \item Domain and error adaptation (+Domain-Adapt., Error-Adapt.)~towards a specific benchmark;
 \item Increasing model depth.
\end{itemize}

Combinations of these generally\footnote{Increasing depth or changing the architecture to the Transformer model is clearly not model-independent.} model-independent methods helped raising the performance of pure neural GEC systems by more than 10\% M$^2$ on the CoNLL 2014 benchmark, also outperforming the previous state-of-the-art \cite{chollampatt2017connecting}, a hybrid phrase-based system with a complex spell-checking system by 2\%. We also showed that a pure neural system can easily outperform a strong pure phrase-based SMT system \cite{junczys2016phrase} when similarly adapted to the GEC task.

On the JFLEG benchmark we outperform the previously-best pure neural system \cite{sakaguchi2017grammatical} by 5.9\% GLEU (4.5\% if no monolingual data is used). Improvements over SMT-based system like \newcite{napoles2017systematically}\footnote{Results based on errata from \url{https://github.com/cnap/smt-for-gec\#errata}} and \newcite{chollampatt2017connecting} are significant and constitute the new state-of-the-art on the JFLEG test set.

\section*{Acknowledgments}
This work was partially funded by Facebook. The views and conclusions contained herein are those of the authors and should not be interpreted as necessarily representing the official policies or endorsements, either expressed or implied, of Facebook.

\bibliography{gecnmt}
\bibliographystyle{acl_natbib}

\end{document}